# Neural Networks with Complex and Quaternion Inputs

Adityan Rishiyur

## 1 Introduction

Many neural network architectures operate only on real data and simple complex inputs. But there are applications where considerations of complex and quaternion inputs are quite desirable. Prior complex neural network models have generalized the Hopfield model, backpropagation and the perceptron learning rule to handle complex inputs. The Hopfield model for inputs and outputs falling on the unit circle in the complex plane was formulated by Noest [15, 16, 17]. Georgiou [2, 3, 12, 14] described the complex perceptron learning rule and the complex domain backpropagation algorithm. Li, Liao and Yu [13] used digital filter theory to perform fast training of complex-valued recurrent neural networks. Recent work by Rajagopal [21] has used Kak's instantaneously trained networks to handle complex inputs.

The instantaneously trained neural network approach [5, 6, 7, 8, 9] utilizes prescriptive learning that requires that the training binary data samples be presented only once during training. In this procedure the network interconnection weights are assigned based entirely on the inputs without any computation. The general method is based on two main ideas that enable the learning and generalization of inputs:

1. The training vectors are mapped to the corners of a multidimensional cube. Each corner is isolated and associated with a neuron in the hidden layer of the network. The outputs of these hidden neurons are combined to produce the target output.

2. Generalization using the radius of generalization enables the classification of any input vector within a Hamming Distance from a stored vector as belonging to the same class as the stored vector.

Tang and Kak [20] present a generalization that is only near-instantaneous, which can handle non-binary inputs. In time-series prediction, a few samples from the series are used for training, and then the network can predict future values in the series. This paper investigates instantaneously trained networks that can handle complex and quaternion inputs. The motivation for examining quaternion inputs is that they permit the consideration of a situation that is midway between classical and quantum systems [24, 25, 26, 27].

## 2 Instantaneously Trained Neural Networks with Complex Inputs

The most popular implementation of Kak's instantaneously trained neural network [5, 6, 7, 8, 9, 21] is the Corner Classification (CC4) algorithm. The CC4 has a minimum of three layers in its architecture: input layer, hidden layer and output layer. The CC4 has a connection from each neuron in the one layer to each neuron in the adjacent layer, so connectivity between layers is high.

The general architecture of the CC4 is shown below. The outputs and inputs of the CC4 can only be 1's and 0's. It can be made unary or binary according to the application for which it is being developed. Unary encoding is preferred over binary encoding. The number of input neurons is one more than the number of input elements in a training sample. This is due to the addition of an extra neuron called the bias neuron that receives a constant input (bias) of 1. The hidden layer constitutes of n neurons. This is because each training vector is associated with (or memorized by) one hidden neuron. The output layer is fully connected, that is all the hidden neurons are connected to each and every output neuron.

Assigning the weights to the links between layers is called training of the neural network. The points are fed in as unary values in sets of four points, shifting one point at a time. This is done till all points are processed. This is the process of assigning weights using the prescriptive learning procedure.

The weights to the links between the neurons are assigned on a rule basis. If the input to the neuron is a 1, the weight assigned is +1 and when the input is a 0, the weight assigned is –1. This applies to all the three layers. We have the bias neuron in the input layer. This weight from this neuron to the neurons in the hidden layer is always r – s + 1, where s is the number of 1's in the input vector and r, as explained earlier, is the radius of generalization for the CC4. In this way the training of the vectors will be done using the training vectors only once. These training vectors are not required during the rest of the prediction process.

The 3C algorithm (from Complex Corner Classification, CCC) is a generalization of the CC4 algorithm. It is capable of training three-layered feedforward networks to map inputs from the complex alphabets {0, 1, i, 1+i} to the real binary outputs 0 and 1
The features of the 3C network are [21]:
    1. Similar to the CC4, the number of input neurons is one more than the number of input elements in a training sample. The extra neuron is the bias neuron which is always set to one.
    2. As in CC4, a hidden neuron is created for all training samples; the first hidden neuron corresponding to the first training sample, the second neuron to the second training sample and so on.
    3. The output layer is fully connected that is each hidden neuron is connected to all the output neurons.
    4. The input alphabets {0, 1, i, 1+i} are treated as complex.
    5. The interconnection weights from all the input neurons excluding the bias neuron are complex.
    6. The weight from the bias neuron to a hidden neuron is assigned as $r - s + 1$, where $r$ is the radius of generalization. The value of $s$ is assigned almost similar to the CC4.

In 3C it is the sum of the number of ones, *i*'s, and twice the number of *(1+i)*'s in the training vector corresponding to the hidden neuron.
7. If the output is 0, the output layer weight is set as -1. If the output is 1, then the weight is set as 1.
8. Due to the combination procedure of the inputs and the weights, the hidden neuron inputs are entirely real.
9. A binary step activation function is used as the activation function at the hidden layer is as well as the output layer.

For complex input neural networks, the two dimensional nature of complex numbers is utilized for encoding. This scheme of encoding called quaternary encoding reduces the network size [21].

## 3 Pattern Classification and Understanding

We describe a pattern classification experiment to analyze the generalization capability of the training algorithm r. The algorithm is used to train a network to separate two regions of any given pattern. The patterns consist of two regions, dividing a 16 by 16 area into a black region and another white region. A point in the black spiral region is represented as a binary "1" and a point in the white region is represented by a binary "0". Any point in the region is represented by row and column coordinates. These coordinates, which are row and column numbers, are encoded and fed as inputs to the network. Quaternary encoding scheme is used for encoding the inputs. To represent a point in the pattern, the 5 - bit strings of the row and column coordinates are concatenated together. The bias is added to this 10 – bit string. The resulting 11-element vector is given as input to the network. The corresponding outputs are 1 or 0, to denote the region that the point belongs to. S

Randomly selected points from both the white and black regions of the pattern are used as training samples. A total of 75 points are used for training. Thus the network used for this pattern classification experiment has 11 neurons in the input layer and 75 neurons in the hidden layer – one hidden neuron for each of the training sample.
After the training is done the network is tested for all 256 points in the 16 by 16 area of the pattern as the value of r is varied from 1 to 4.

### 3.1 Performance Analysis

Even though the complex input neural network provides good classification, it is found that as the value of the radius of generalization, r is increased, the black region increases in size as well. To determine the amount of error in the classification, the number of points classified and misclassified in the four experiments is calculated. From this data, an error coefficient is calculated which is the inherent classification error coefficient of the complex input neural network for a particular value of n - the number of input coordinates used in the training of the network. The efficiency of the algorithm for the increasing values of r is also calculated from the knowledge of classified points in the patterns.

The error coefficient is calculated as a ratio of the number of misclassified points to the total number of points in the pattern. The total number of points in the patterns that have been used for our analysis is 256. The efficiency of the algorithm is calculated as a ratio of the number of points that were correctly classified to the total number of points in the pattern and is expressed as a percentage.

A summary of the experiments is presented for the spiral, rectangular box, star and man-like patterns respectively. These tables contain the number of points that were correctly classified and misclassified by the algorithm, the error coefficient of each case and the efficiency of the algorithm for varying values of radius of generalization. This is the classification error for the patterns when the number of training samples is 75.

When the observations are analyzed, it is clear that for a constant value of n, as the radius of generalization increases, the area of black region increases in most cases and it leads to an increase in the error of classification. The efficiency of the algorithm correspondingly decreases. We can sufficiently conclude that lower values of r provide better classification given that the number of training samples is large enough.

The previous analysis calculated the error when the value of n was held constant. For further analysis on the relationship between error in classification and the number of training samples n, different set of pattern classification experiments are conducted. The number of training samples, n is gradually decreased and the performance of the algorithm in classifying the points is observed for varying degrees of radius of generalization. The classification error of the network for each case of n is calculated from the data.

From the above data, the relationship between the number of training samples n, the radius of generalization r, and the classification error E is analyzed. The classification error for each of the cases is presented in Table 3.1.

**Table 3.1: Classification error for different values of n**

|       | n = 75 | n = 65 | n = 55 | n = 35 | n = 25 |
|-------|--------|--------|--------|--------|--------|
| r = 1 | 0.129  | 0.145  | 0.156  | 0.219  | 0.246  |
| r = 2 | 0.137  | 0.125  | 0.16   | 0.219  | 0.211  |
| r = 3 | 0.184  | 0.172  | 0.199  | 0.254  | 0.273  |
| r = 4 | 0.258  | 0.254  | 0.238  | 0.309  | 0.355  |

It is observed that for n = 65, better classification is obtained when radius of generalization is 2. For values of n = 55 and n = 45, the classification error is almost the same for r=1 and r=2. For an even lower number of samples, n = 25, r = 2 provides a much better performance in classification. From this data and the graphs we can conclude that when the number of training samples is reduced, better classification is achieved by the network if the radius of generalization is 2.

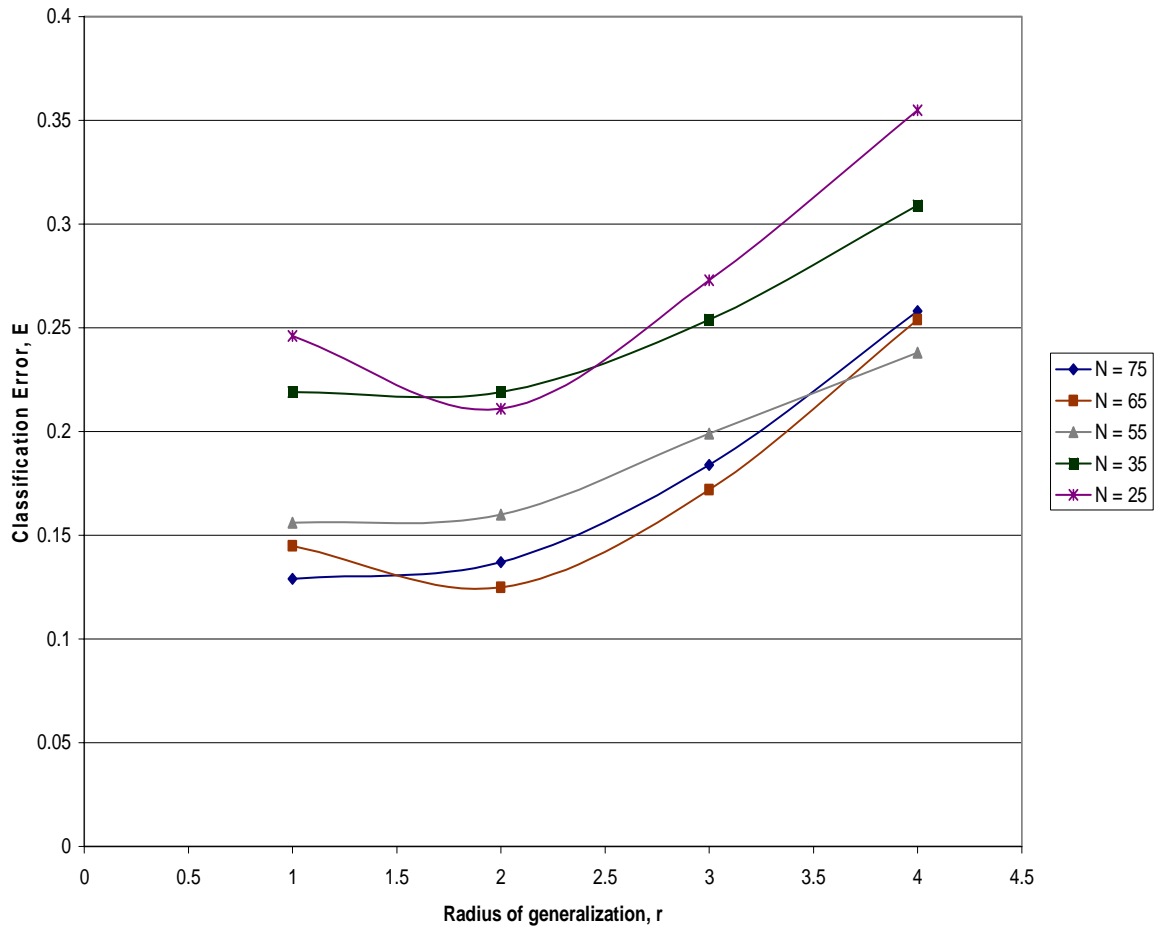

**Figure 3.1: Consolidated performance curve for different values of n**

It is desirable to find out the optimum number of training samples needed for any general pattern for most efficient classificatory performance of the network. The optimum value of the ratio of training samples n to the total number of points in a pattern N for each value of radius of generalization is determined.

From the graphs it can be deduced that classification error for a given pattern is a minimum when the ratio of training sample, n and total number of points, N reaches an optimum value. This value varies between 0.20 and 0.30 depending on the value of r. As the value of r decreases, the optimal n/N ratio for minimal classification error increases, that is for a lower value of r, more samples are required for better classification.

It has been suggested that instantaneously trained neural networks are a good model for short-term memory and learning [9]. Given that, our experimental data leads us to propose that human image understanding uses a two-step approach where the image is first seen in its skeleton using r=0. The filling in, which represents the generalization from this skeleton, is done by considering other patterns that are stored in the human memory system. Since this filling in does not involve the scanning of the physical image, it is very fast. Also since this filling in will vary across subjects, this explains why perception as well as recall have a subjective element.

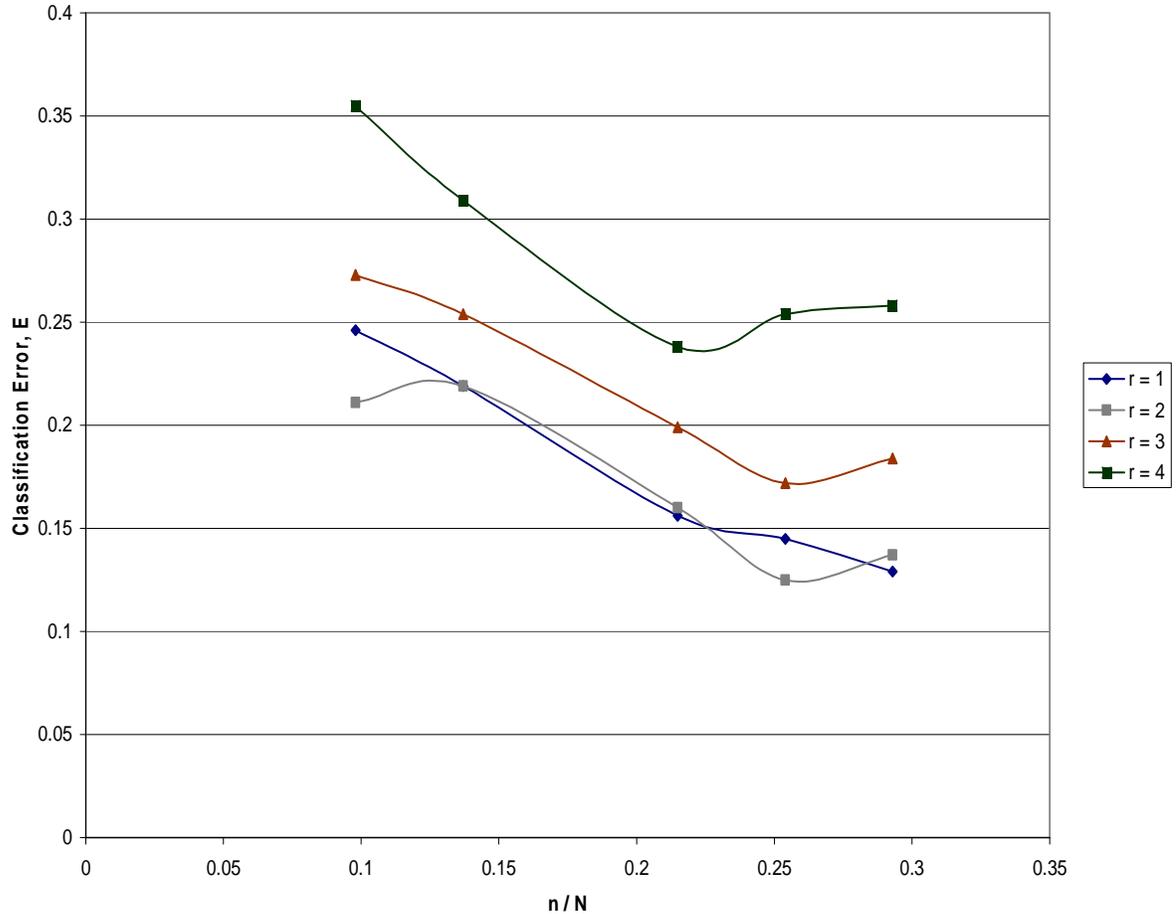

**Figure 3.2: Consolidated performance curve for different values of r**

This seems to be supported by psychological experiments that show that the recollection of well-defined patterns (such as the image on a coin) varies from individual to individual. This is one of the significant insights to come out of the present investigation. We do not know if this insight can form the basis of a new approach to machine understanding of images. But it appears that this is a good model to study biological vision.

**3.2 Complexity and Capacity**

The network that we investigated in the previous section used integer values of the data. This means that one must use a very large network to deal with the number of input levels that can be distinguished. If we could consider real-input networks then the complexity of the network will be reduced dramatically. This has been accomplished, without excessive cost in terms of additional processing and lost speed, for non-complex neural networks [20]. The generalization of these modified networks to complex inputs increased their computation capacity and range of applications.

## 4 Quaternions and Their Applications

Quaternions are a non-commutative extension of the complex numbers. Quaternions find uses in both theoretical and applied mathematics, in particular for calculations involving multi-dimensional rotations. They extend the concept of rotation in three dimensions to rotation in four dimensions. In effect, they may be considered to add an additional rotation angle to spherical coordinates i.e. Longitude, Latitude and Rotation angles. The algebra of quaternions is often denoted by **H.** While the quaternions are not commutative, they are associative, and they form a group known as the quaternion group. While the complex numbers are obtained by adding the element *i* to the real numbers which satisfies $i^2 = -1$, the quaternions are obtained by adding the elements i, j and k to the real numbers which satisfy the following relations:

$$i^2 = j^2 = k^2 = ijk = -1$$

If the multiplication is assumed to be associative (as indeed it is), the following relations follow directly:

$$ij = k, \qquad ji = -k,$$
$$jk = i, \qquad kj = -i,$$
$$ki = j, \qquad ik = -j,$$

Every quaternion is a real linear combination of the basis quaternions *1, i, j*, and *k*, i.e. every quaternion is uniquely expressible in the form

$$H = a.1 + b\,i + c\,j + d\,k$$

where *a, b, c,* and *d* are real numbers. In other words, as a vector space over the real numbers, the set H of all quaternions has dimension 4, whereas the complex number plane has dimension 2. Addition of quaternions is accomplished by adding corresponding coefficients, as with the complex numbers. By linearity, multiplication of quaternions is completely determined by the multiplication table above for the basis quaternions. Under this multiplication, the basis quaternions, with their negatives, form the quaternion group of order 8, Q8. The scalar part of the quaternion is a, whereas the remainder is the vector part. Thus a vector in the context of quaternions has zero for scalar part.

The *conjugate* $z^*$ of the quaternion $z = a + bi + cj + dk$ is defined as $z^* = a - bi - cj - dk$ and the *absolute value* of z is the non-negative real number defined by $|z| = \sqrt{zz^*} = \sqrt{a^2 + b^2 + c^2 + d^2}$.

### 4.1 Quaternion-based Systems

Quaternions find various applications in classical mechanics, quantum mechanics, and the theory of relativity. Quaternions have been used in aerospace applications and flight simulators, particularly when inertial attitude referencing and related control schemes are employed. More recently, graphics and game programmers discovered the potential of quaternions and started using it as a powerful tool for describing rotations about an arbitrary axis. From computer graphics, the application domain of quaternions has expanded into other fields such as visualization, fractals and virtual reality [22].

A model of computation based on quaternions, which is inspired on the quantum computing model, has been proposed. Pure states are vectors of a suitable linear space over the quaternions. Other aspects of the theory are the same as in quantum computing: superposition and linearity of the state space, unitarity of the transformations, and projective measurements. The quaternionic equivalent to the qubit, as the most elementary quaternionic information measure, is called a quaterbit [24].

## 5 Quaternion Encoding Scheme

For implementing the complex input neural network, quaternary encoding was used. This scheme took advantage of the 2-dimensional properties of complex numbers. By using this technique, the number of inputs was reduced from 33 (16 x-coordinate + 16 y-coordinate + 1 bias) in the unary encoding scheme to 11 in this new scheme. It was a modification of the unary scheme and introduced two additional characters i and 1+i. By using quaternions, which are extensions of complex numbers, it is possible to reduce the network size even further. But this requires the development of a suitable encoding scheme. Our proposed encoding scheme is called quaternion encoding. It accommodates additional characters j, k, 1+j, 1+k, i+j, j+k, i+k, 1+i+j, 1+j+k, 1+i+k, i+j+k and 1+i+j+k to the existing set of inputs. The row and column indices that range from 1 to 16 can now be represented by quaternion *codewords*. Table 6.1 holds all the codewords used to represent the integers 1 to 16.

### 5.1 Length of Codewords

An important issue is to decide the length of the codewords that would be required to represent a given range of integers. Let $l$ be the length of the codewords for a range of $C$ integers. For the range of integers in the Table 6.1, $C = 16$ and $l = 1$. The relationship between $l$ and $C$ is deduced to be

$$l = \textbf{ceil} \, [(C - 1) / 15] \qquad [5.1]$$

When (C-1) is not divisible by 15, more than the required number of codewords is formed. In such a case, any $C$ consecutive codewords may be used from the complete set.

**Table 5.1: Quaternion codewords for integers 1 to 16**

| Integer | Quaternion code |
|---|---|
| 1 | 0 |
| 2 | 1 |
| 3 | i |
| 4 | j |
| 5 | k |
| 6 | 1+i |
| 7 | 1+j |
| 8 | 1+k |
| 9 | i+j |
| 10 | j+k |
| 11 | i+k |
| 12 | 1+i+j |
| 13 | 1+j+k |
| 14 | 1+i+k |
| 15 | i+j+k |
| 16 | 1+i+j+k |

## **Example 5.1**

If we need to represent a range of 31 integers, we require 31 codewords. We can calculate *l* by using equation 6.1. Here $C = 31$

$l =$ **ceil** $[(31 - 1) / 15]$
$l =$ **ceil** $[30 / 15] = 2$

**Table 5.2 All possible quaternion codewords for $l = 2$**

| Integer | Quaternion Code | |
|---|---|---|
| 1 | 0 | 0 |
| 2 | 0 | 1 |
| 3 | 1 | 1 |
| 4 | 1 | i |
| 5 | i | i |
| 6 | i | j |
| 7 | j | j |
| 8 | j | k |
| 9 | k | k |
| 10 | k | 1+i |
| 11 | 1+i | 1+i |
| 12 | 1+i | 1+j |
| 13 | 1+j | 1+j |
| 14 | 1+j | 1+k |
| 15 | 1+k | 1+k |
| 16 | 1+k | i+j |
| 17 | i+j | i+j |
| 18 | i+j | j+k |
| 19 | j+k | j+k |
| 20 | j+k | i+k |
| 21 | i+k | i+k |
| 22 | i+k | 1+i+j |
| 23 | 1+i+j | 1+i+j |
| 24 | 1+i+j | 1+j+k |
| 25 | 1+j+k | 1+j+k |
| 26 | 1+j+k | 1+i+k |
| 27 | 1+i+k | 1+i+k |
| 28 | 1+i+k | i+j+k |
| 29 | i+j+k | i+j+k |
| 30 | i+j+k | 1+i+j+k |
| 31 | 1+i+j+k | 1+i+j+k |

From table 5.2 we can observe that 31 codewords can be formed with $l = 2$.

In a similar way when $C = 36$, we can deduce from equation 6.1 $l = 3$. Using $l = 3$, 46 different codewords can be constructed of which any 36 can be used for the encoding.

**5.2 Quaternion Neural Networks**

Quaternion neural networks (QNN) are models in which computations of the neurons are based on quaternions, the four-dimensional equivalents of imaginary numbers. Since quaternion numbers are non-commutative on multiplication, different models can be considered.

Quaternion representations are more efficient that that of matrices and operations on them such as composition can be computed more efficiently. Quaternions are also applied in control theory, signal processing, attitude control, physics, and orbital mechanics, mainly for representing rotations/orientations in three dimensions. For example, it is common for spacecraft attitude-control systems to be commanded in terms of quaternions, which are also used to telemeter their current attitude. The rationale is that combining many quaternion transformations is more numerically stable than combining many matrix transformations, avoiding such phenomena as gimbal lock, which can occur when Euler angles are used. Using quaternions also reduces overhead from that when rotation matrices are used, because one carries only four components, not nine, the multiplication algorithms to combine successive rotations are faster, and it is far easier to renormalize the result afterwards. Backpropagation based QNNs suffer from slow training.

**5.3 QNNs Based on Instantaneous Training**

We have examined QNNs that generalize instantaneously trained complex neural networks. Since their real advantage is speed, they can be put to use in real-time applications such as robotics, visualization, and graphics. This should allow one to use quaternions in a manner that is much more flexible than in the current rule-based systems. The requirement of the large number of hidden neurons might not be too much of a cost in many applications.

Although backpropagation networks have been investigated for their use in biomechanical systems, it is difficult to see how they could actually form the basis of such function for they learn very slowly. A credible model for biological function must have a component that is very efficient, and we believe that it must be able to learn instantaneously or near-instantaneously. In other words, just as biological memory comes in two varieties: short-term, which is near-instantaneous, and long-term, which is relatively slow, three-dimensional control in biological systems must also have one component that acts very quickly.

We conjecture that biological systems use an instantaneously trained neural network based on quaternions or even higher-dimensional entities (such an octonions or quaternion-based higher structures), that can deal with non-integer values, for their three-dimensional movements. Physiological experiments would be required to investigate this question further.

# 6 Conclusion

This article has investigated the use of instantaneously trained neural networks with complex and quaternion inputs, based on the corner classification approach, motivated by possible applications in robotics, visualization and computer graphics. We have analyzed the performance of the basic algorithm and shown how it provides a plausible model of human perception and understanding of images. We have also suggested ways to extend the algorithm into the quaternion space.

Encoding techniques for the input data were discussed. A new encoding technique with quaternion inputs was proposed as an extension to the quaternary encoding. Quaternion encoding greatly reduces the size of the network in comparison to the earlier encoding schemes. A benefit of this encoding is that the number of input neurons required will be greatly reduced when compared to the earlier networks.

As future work, we suggest further study of quaternion and octonion input instantaneously trained neural networks. It would be useful to investigate how the hidden neurons in such networks could be reduced in number, without a corresponding loss in generalization function, so as to lessen memory requirement in their implementation. Another problem worthy of further research is to investigate complex and quaternion models that are different from the ones that form the basis of our work. Specifically, our model is limited to integer values, and it needs to be generalized so as to be able to deal with real values. Such a generalized model might have the computational power to explain the manipulation capacity of biological systems.